\documentclass[letterpaper]{article} 
\usepackage{aaai23}  
\usepackage{times}  
\usepackage{helvet}  
\usepackage{courier}  
\usepackage[hyphens]{url}  
\usepackage{graphicx} 
\usepackage{natbib}  
\usepackage{caption} 
\usepackage{algorithm}
\usepackage{algorithmic}

\usepackage{booktabs}
\usepackage{multirow}
\usepackage{multicol}
\usepackage{diagbox}
\usepackage{amssymb}

\usepackage{newfloat}
\usepackage{amsmath}
\usepackage{listings}
\DeclareCaptionStyle{ruled}{labelfont=normalfont,labelsep=colon,strut=off} 
\floatstyle{ruled}
\newfloat{listing}{tb}{lst}{}
\floatname{listing}{Listing}
\newcommand{\ie}{\textit{i}.\textit{e}.}

\newcommand{\etal}{\textit{et al.}}

\title{Contrastive Predictive Autoencoders for Dynamic Point Cloud \\ Self-Supervised Learning}
\author{
    Xiaoxiao Sheng\textsuperscript{\rm }\equalcontrib,
    Zhiqiang Shen\textsuperscript{\rm }\equalcontrib,
    Gang Xiao\textsuperscript{\rm }\thanks{Gang Xiao is the corresponding author.}
}
\affiliations{
    \textsuperscript{\rm} Shanghai Jiao Tong University\\

    $\{$shengxiaoxiao, shenzhiqiang, xiaogang$\}$@sjtu.edu.cn
}

\usepackage{bibentry}

\begin{document}

\maketitle

\begin{abstract}
We present a new self-supervised paradigm on point cloud sequence understanding.
Inspired by the discriminative and generative self-supervised methods, we design two tasks, namely point cloud sequence based  Contrastive Prediction and Reconstruction (CPR), to collaboratively learn more comprehensive spatiotemporal representations. 
Specifically, dense point cloud segments are first input into an encoder to extract embeddings. 
All but the last ones are then aggregated by a context-aware autoregressor to make predictions for the last target segment. Towards the goal of modeling multi-granularity structures, local and global contrastive learning are performed between predictions and targets. 
To further improve the generalization of representations, the predictions are also utilized to reconstruct raw point cloud sequences by a decoder, where point cloud colorization is employed to discriminate against different frames. 
By combining classic contrast and reconstruction paradigms, it makes the learned representations with both global discrimination and local perception. 
We conduct experiments on four point cloud sequence benchmarks, and report the results on action recognition and gesture recognition under multiple experimental settings. 
The performances are comparable with supervised methods and show powerful transferability.

\end{abstract}

\section{Introduction}

\noindent 

The development of depth sensors facilitates the acquisition of dynamic point clouds and makes it widely applied in many scenarios, such as autonomous vehicles and robots.
Currently, processing these real-time point clouds for surroundings perception still relies on supervised methods.
However, labeling plenty of dynamic point clouds is quite labor-intensive and error-prone.
Meanwhile, self-supervised methods in images make great success, significantly alleviating annotation requirement and even outperforming supervised pretraining methods in many tasks \cite{zbontar2021twins,hemocov1}.
Inspired by this, this paper focuses on learning dynamic point cloud representations in a self-supervised manner.

\begin{figure}[t]
	\centering
	\includegraphics[width=1\columnwidth]{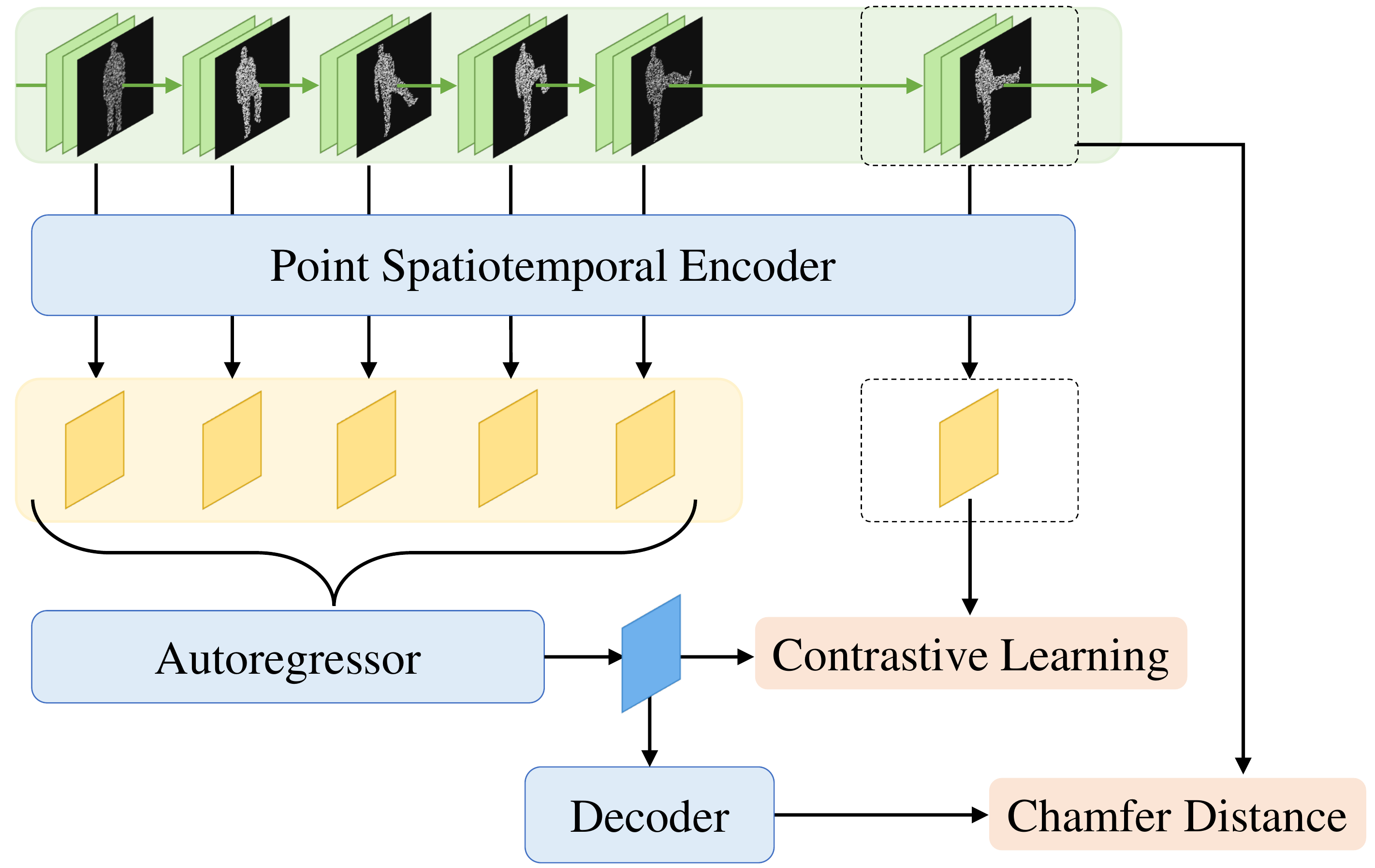}
	\caption{Illustration of our main idea. We combine contrastive predictive coding and reconstruction to establish a unified self-supervised framework for dynamic point clouds. }

	\label{motivation}
\end{figure}

Classical self-supervised methods in images always rely on contrasting the samples with strong data augmentations or reconstructing the randomly masked patches based on visible parts to obtain high-level semantics \cite{zbontar2021twins,bao2021beit}.
For discriminative methods, rich and diverse sample pairs are usually created.
However, it is likely that the semantic consistency of the augmented dynamic point cloud pairs is difficult to be guaranteed.
Furthermore, experimental results in \cite{xie2020pointcontrast} also demonstrate that real multi-camera views are more effective than hand-crafted views for point clouds.
For generative self-supervised methods, the original point coordinates need to be recovered.
However, the corresponding coordinates will be added to masked tokens as positional embeddings.
This results in information leakage and makes the task less challenging.
Based on the above analysis, it is natural to ask a question: \emph{how can we leverage the advantages of discriminative and generative methods for self-supervised learning on dynamic point clouds?}

Considering that point cloud sequences are constantly evolving with geometric dynamics, we can explore temporal-related self-supervised signals lying in the sequence itself, rather than relying on massive augmented data. Contrastive predictive coding is to predict about future based on recent past of the sequences, which has been successfully validated in speeches and videos \cite{oord2018cpcv1,han2020memory}. Generally, this paradigm aims to use powerful encoder and autoregressor to learn high-level representations with a contrastive pretext task.
In addition, to introduce more task-independent properties, we explore to maximize the mutual information between representations and inputs.
Therefore, we integrate point cloud sequence reconstruction into contrastive predictive coding for learning more generalized representations.
Our main idea is illustrated in Figure \ref{motivation}. 

Specifically, we first sample multiple segments and then send them into an encoder to extract spatiotemporal features.
The features of all but the last segments are regarded as tokens to fed into a transformer autoregressor to predict future states.
We perform contrastive tasks between the predictions and target segment features.
However, due to the unordered and irregular properties of point cloud, the points attached with predicted spatiotemporal features are not aligned to those target ones.
We design an interpolation based method to achieve points alignment and further update predicted features correspondingly.
Then, local contrastive learning is performed, and we also explore hard negatives to enhance the discriminability of the model.
Meanwhile, global contrastive learning is conducted to compare the embeddings of class tokens with sequence-level features of entire current and predicted segments.
In addition, raw point cloud sequences of the target segment are reconstructed based on the predictions.
Point cloud colorization is applied to the target segment for further discriminating frames.
Overall, contrastive prediction task explores more on the discrimination between diverse point cloud sequences, while generative self-supervised task perceives internal structures of point cloud itself.
By combining these two self-supervised tasks adapted to dynamic point cloud, it enables the learned representations more comprehensive with both powerful instance discriminability and regional context-awareness.

Multi-granularity representations can be learned by establishing our unified contrastive prediction and reconstruction (CPR) self-supervised framework.
We evaluate our method on four dynamic point cloud benchmarks, including MSRAction3D \cite{msr}, NTU-RGBD~60 \cite{ntu60}, NvGesture \cite{molchanov2016nvg}, and SHREC'17 \cite{de2017shrec}.
Our method achieves excellent performance comparable with state-of-the-art supervised methods.
Ablation studies are performed to investigate the design of self-supervised tasks.
The main contributions of this paper are as follows:

\begin{itemize}
\item We propose a new contrastive prediction and reconstruction self-supervised framework for dynamic point cloud understanding. 

\item We conduct local and global contrastive predictions to learn comprehensive representations with both context-awareness and holistic discrimination.

\item We reconstruct raw point cloud sequences to promote the generalization of learned representations, and apply point cloud colorization to discriminate different frames.

\item We perform extensive experiments on multiple dynamic point cloud benchmarks. We also show detailed ablation studies to analyze how to design self-supervised tasks. 

\end{itemize}

\section{Related Work}

\subsection{Self-supervised Methods for Images}
Benefiting from massive data, self-supervised methods in images achieve remarkable performance.
This not only alleviates the demands for labeling, but further reveals that more generalized representations are obtained in a self-supervised manner without introducing label ambiguity.
Discriminative self-supervised methods compare two views of one sample processed by diverse data augmentations to learn high-level semantics, and utilize plenty of negative samples for contrast to promote instance discriminability \cite{chen2020simclrv1}.
More effective techniques are also explored to enhance the generalization of pretrained encoder for downstream tasks, such as dynamic queues \cite{hemocov1}, momentum updating \cite{hemocov1}, stop-gradients \cite{chen2021simsiam}, and clustering \cite{caron2020swav}.
Generative self-supervised methods aim to learn the generalized representation by mask image modeling \cite{bao2021beit,xie2022simmim,he2022mae}.
The masked and unmasked tokens can be simultaneously processed by an encoder, and an extra decoder is necessary to reconstruct raw input or low-level features based on the latent representations \cite{bao2021beit}.
An asymmetric network structure is also explored to only process the visible patches by the encoder, which significantly improves computation efficiency \cite{he2022mae}.

\subsection{Self-supervised Methods for Videos}
Self-supervised methods of videos further consider the properties of sequence data itself, such as temporal continuity and redundancy. Similarly, discriminative methods focus on how to build sample pairs from diverse video clips \cite{videomoco,qian2021spatiotemporal}. 
Generative methods randomly mask the spatiotemporal tubes with a higher ratio to prevent information leakage \cite{tong2022videomae,wang2022bevt}.
Besides, many works have been proposed to model motions, such as co-training of two models based RGB and optical flow separately \cite{han2020selfcotrain}, or building decoupled static and dynamic concepts \cite{qian2022static}.
Clearly, contrastive learning based methods emphasize the discrimination between diverse samples, while reconstruction based methods focus on the perception and understanding of samples themselves.
They are both limited in learning multi-granularity representations.

\subsection{Self-supervised Methods for Static Point Clouds}
Inspired by the success of self-supervised methods in images, many works focus on learning point cloud representations in a self-supervised manner.
PointContrast \cite{xie2020pointcontrast} is proposed to contrast real-world multi-view point clouds with various data augmentations.
Unfortunately, multi-camera viewpoints are not available in some scenarios.
As a BERT-style self-supervised framework, Point-BERT \cite{yu2022pointbert} predicts latent representations learned by an offline tokenizer, which causes two-stage pretraining.
In addition, directly recovering raw point cloud easily leads to information leakage due to the positional encoding of masked tokens.
To alleviate the above problem, PointMAE \cite{pang2022pointmae} is proposed to only input unmasked tokens into encoder and add the masked tokens to the decoder for reconstructing.
Alternatively, MaskPoint \cite{liu2022pointdiscrimination} is proposed to randomly select visible points into encoder and train a decoder to discriminate between masked points and noises points. It combines the discriminative method and the masking task to achieve excellent performance.

\subsection{Modeling Dynamic Point Clouds}
Currently, modeling dynamic point cloud is still dominated by traditional supervised methods \cite{zhong2022nopain,fan2021deep,PST2,action4d,wen2022point,fan2022point}.
MeteorNet \cite{MeteorNet} gradually learns aggregated features for each point by constructing point-wise spatiotemporal neighborhoods.
3DV \cite{3dv} encodes motion information utilizing regular voxels and then abstracts these dynamic voxels into point sets to model spatiotemporal features.
PSTNet \cite{pstnet} hierarchically extracts features of raw point cloud sequences with spatial and temporal decoupled convolutions.
P4Transformer \cite{p4d} first utilizes the point tube convolution to build tokens and further aggregates these tokens by a transformer encoder.
Wang~\etal~\cite{wang2021selfwacv} proposes an order prediction self-supervised task on shuffled point cloud clips to learn dynamic point cloud features.
However, this method only mines temporal information, while ignoring spatiotemporal modeling.
Therefore, the spatiotemporal discriminability and local context awareness of representations learned by this manner are limited.

Unlike the previous methods, we propose a new self-supervised paradigm for modeling point cloud videos.
We combine the advantages of contrastive learning and generative methods to design contrastive prediction and reconstruction tasks for dynamic point clouds, jointly facilitating the generalization performance for downstream tasks.

\begin{figure*}[ht]
        \centering
        \setlength{\abovecaptionskip}{0.4cm}
        \includegraphics[width=0.96\linewidth]{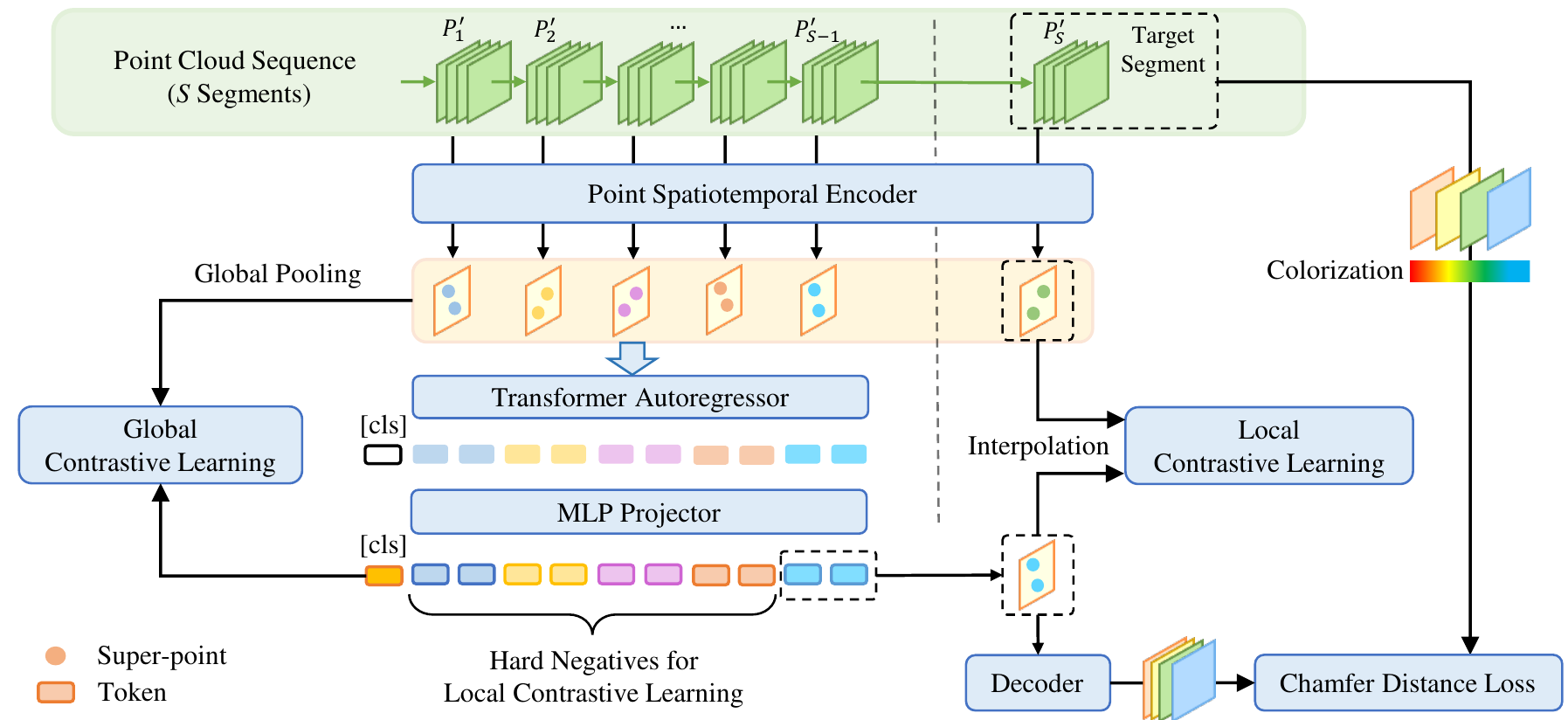}
        \caption{The framework of our approach. It contains three main components, \ie, point spatiotemproal encoder, transformer autoregressor, and decoder. By organically combining these components, we construct contrastive prediction and reconstruction self-supervised tasks.}
        \label{fw}
\end{figure*}

\section{Contrastive Prediction and Reconstruction}

We present the overall framework of our method in Figure~\ref{fw}, which includes three main modules.
\emph{Point Spatiotemporal Encoder} is first utilized to aggregate dense point cloud segments.
Then, we input the aggregated embeddings into a \emph{Transformer Autoregressor} to make predictions for the last target segment.
An MLP head is introduced to further project the representations into latent space for contrast.
We perform local and global contrastive learning between predictions and targets to progressively capture multi-granularity features.
In addition, raw point cloud sequences are reconstructed by a \emph{Decoder} based on the predictions.
In this section, we first briefly introduce the encoder and autoregressor, then present contrastive prediction and reconstruction self-supervised tasks in detail.

\subsection{Point Spatiotemporal Encoder}

A point cloud sequence is denoted as $\boldsymbol{P}\in\mathbb{R}^{T\times N\times3}$, where $T$ is the sequence length and $N$ is point number in each frame.
PSTNet \cite{pstnet} is a typical spatial and temporal decoupled feature extractor for dynamic point clouds.
Specifically, the farthest point sampling is first utilized to sample $N^\prime$ anchor points within each frame. 
Then, anchors are mapped into adjacent frames to construct spatiotemporal point tubes. 
Spatial convolution is built to extract local structures within the neighborhoods grouped by \emph{k}-nearest neighbor algorithm. 
1D convolution is utilized to capture temporal features on these tubes. 
These aggregated spatiotemporal features are denoted as $\{\boldsymbol{x}_i\}_{i=1}^m$, where $m$ is the number of aggregated super-points.
In this paper, we focus more on how to design self-supervised tasks for learning point spatiotemporal representations. 
Therefore, we directly adopt the backbone of PSTNet as our encoder.

\subsection{Transformer Autoregressor}

The autoregressor is to predict future representations based on spatiotemporal features extracted by the encoder.
In classic contrastive prediction based methods \cite{oord2018cpcv1,han2020memory}, LSTMs are always used as autoregressors to predict future states of sequence data. 
However, when dealing with long sequences, LSTMs are prone to catastrophic forgetting, causing inferior predictions.
Considering that 3D action recognition strongly relies on long-term temporal information, we introduce the powerful transformer as the autoregressor.
We adopt the standard transformer which consists of multi-head self-attentions and feed-forward networks.
We treat the above $\{\boldsymbol{x}_i\}_{i=1}^m$ as tokens.
Positional embeddings $\{\boldsymbol{e}_i\}_{i=1}^m$ are obtained by projecting the spatiotemporal coordinates ($x,y,z,t$) of super-points with a linear function.
$\mathbf{E}[\mathrm{s}]$ is the class token and $\boldsymbol{e}_0$ is its positional embedding obtained by projecting $(x_0, y_0, z_0, t_0)$, where ($x_0, y_0, z_0)$ is mean value of target segment points and $t_0$ is the timestamp of target segment.
Finally, the token sequence $[\mathbf{E}[\mathrm{s}]+\boldsymbol{e}_0, \ \boldsymbol{x}_1+\boldsymbol{e}_1,\ \ldots,\ \boldsymbol{x}_m+\boldsymbol{e}_m]$ is sent into the transformer to model global spatiotemporal relations and make predictions.

\subsection{Contrastive Prediction Task}

We divide the whole input sequence $\boldsymbol{P}\in\mathbb{R}^{T\times N\times3}$ into $S$ segments equally.
Each segment is $\boldsymbol{P}^\prime\in\mathbb{R}^{M\times N\times3}$, and $T=S\times M$.
We utilize the point spatiotemporal encoder to obtain target embeddings $\boldsymbol{Z}\in\mathbb{R}^{l \times r \times c}$ of the $S$-th target segment, where $l$ is the frame number after aggregation, $r$ is the super-point number, and $c$ is embedding channel.
The former $S\!\!-\!\!1$ segments are input into the encoder and autoregressor to predict target embeddings. 
Because of temporal adjacency with target segment, the updated embeddings of the $(S\!\!-\!\!1)$th segment are taken as predictions, denoted as $\boldsymbol{Q}\in\mathbb{R}^{l \times r \times c}$.
However, due to the unordered and irregular properties of point clouds, it can not directly align predictions and target embeddings.
To alleviate this dilemma, we interpolate the predicted features using \emph{k}-nearest neighbor algorithm by centering at target super-points, where $k=3$.
We denote the interpolated predictions as $\boldsymbol{\hat Q}\in\mathbb{R}^{l \times r \times c}$. 
The aligned embeddings are regarded as positive pairs ($\boldsymbol{z}_i, \boldsymbol{\hat q}_+$), and the remainings are negative samples. 
Moreover, the embeddings belonging to former $S\!-\!2$ segments are explored as hard negatives to improve discriminability of the model. 
Then, local contrastive learning is conducted to obtain fine-grained features. 
We utilize local Info Noise Contrastive Estimation (InfoNCE) loss as follows:
\begin{small}
\begin{equation}
    \mathcal{L}_{l}\!=\!-\!\sum_{\boldsymbol{z}_i\in\boldsymbol{Z}}\!\log\!\frac{\exp(\boldsymbol{z}_i^T \boldsymbol{\hat q}_+/ \tau)}{\exp(\boldsymbol{z}_i^T \boldsymbol{\hat q}_+/\tau)\!+\!\sum_{\boldsymbol{q}_j\in{\Psi}}{\exp{(\boldsymbol{z}_i^T \boldsymbol{q}_j/\tau)}}},
\end{equation}
\end{small} 

\noindent where $\Psi$ is a set that contains the negatives, and $\tau$ is temperature hyper-parameter. 

To explore holistic semantics, we obtain the global representations of whole input sequence, denoted as $\boldsymbol{H}\in\mathbb{R}^{\mathcal{B}\times c}$, by performing max-pooling on the embeddings of $S$ segments passed through the encoder.
Then, we perform global contrastive learning between the representation of class token and $\boldsymbol{H}$.
The corresponding sample pairs are positives, otherwise negatives.
The global InfoNCE loss is as follows:
\begin{small}
\begin{equation}
    \mathcal{L}_{g}\!=\!-\!\!\sum_{\boldsymbol{h}_i\in\boldsymbol{H}}\!\log\!\frac{\exp(\boldsymbol{h}_i^T \boldsymbol{\hat g}_+/ \tau)}{\exp(\boldsymbol{h}_i^T \boldsymbol{\hat g}_+/\tau)\!+\!\sum_{\boldsymbol{g}_j\in{\Theta}}{\exp{(\boldsymbol{h}_i^T \boldsymbol{g}_j/\tau)}}},
\end{equation}
\end{small} 

\noindent where $\boldsymbol{h}_i$ represents the embedding of $i$-th input sequence, $\boldsymbol{\hat g}_+$ is the embedding of its class token, and $\Theta$ contains all negatives. 

By combining local and global contrastive learning, our method effectively captures multi-granularity features. Moreover, we explore more meaningful hard negatives and present their effectiveness in ablation studies.

\subsection{Point Cloud Sequences Reconstruction Task}

Theoretical analysis in \cite{analysis} points out that reconstructing raw inputs maximizes mutual information between representations and inputs to promote generalizability.
Inspired by this, we design sequences reconstruction using the predictions.
Rather than recovering all points of target segment, we perform spatial downsampling by the farthest point sampling.
This avoids paying more attention to low-level details and reducing the computational burden. 
Specifically, we colorize each frame of downsampled target segment with specific RGB values to distinguish different frames. 
With the increase of the temporal index, the corresponding color changes from red to green to blue \cite{color}. 
We denote the colorized target points as $\boldsymbol{\hat P}_t^\prime\in\mathbb{R}^{M\times N^\prime \times6}$, where $N^\prime$ is the number of downsampled points in each frame and 6 means $(x,y,z,r,g,b)$.

\begin{table*}[t]
    \centering
    \begin{tabular}{l|c|ccccc}
    \toprule
    \textbf{Methods} & \textbf{Input} & \multicolumn{5}{c}{\textbf{Accuracy}}\\
    \midrule
    Vieira~\etal~\cite{vieira2012stop}   & depth map       & \multicolumn{5}{c}{78.20 \ \ \ (20 frames)}\\
    Kl$\ddot{a}$ser\etal~\cite{klaser}  & depth map    & \multicolumn{5}{c}{81.43 \ \ \ (18 frames)}\\
    Actionlet~\cite{miningactionlet}          & skeleton     & \multicolumn{5}{c}{88.21 \ \ \ (all frames)}\\
    \midrule
    \midrule
    \multicolumn{2}{c}{\textbf{Frames:}} & 4     & 8     & 12    & 16    & 24 \\
    \midrule
    MeteorNet~\cite{MeteorNet}           & point         & 78.11 & 81.14 & 86.53 & 88.21 & 88.50 \\
    PSTNet~\cite{pstnet}                 & point         & 81.14 & 83.50 & 87.88 & 89.90 & 91.20 \\
    P4Transformer~\cite{p4d}             & point         & 80.13 & 83.17 & 87.54 & 89.56 & 90.94 \\
    PSTNet++~\cite{fan2021deep}          & point         & 81.53 & 83.50 & 88.15 & 90.24 & 92.68 \\
    PST-Transformer~\cite{fan2022point}  & point         & 81.14 & 83.97 & 88.15 & 91.98 & \textbf{93.73}\\
    PST$^2$~\cite{PST2}                  & point         & 81.14 & 86.53 & 88.55 & 89.22 & - \\
    Kinet~\cite{zhong2022nopain}             & point         & 79.80 & 83.84 & 88.53 & 91.92 & 93.27 \\
    PPTr~\cite{wen2022point}             & point         & 80.97 & 84.02 & 89.89 & 90.31 & 92.33 \\
    \midrule
    4D MinkNet + ROP Pretraining~\cite{wang2021selfwacv} & point         & - & 86.31 & - & - & -\\
    MeteorNet + ROP Pretraining~\cite{wang2021selfwacv}  & point         & - & 85.40 & - & - & -\\
    \midrule
    \textbf{CPR~(Ours)} & point & \textbf{82.50} & \textbf{86.53} & \textbf{91.00} & \textbf{92.15} & 93.03 \\
    \bottomrule
    \end{tabular}
    \caption{Action recognition accuracy (\%) on the MSRAction3D dataset.}
    \label{MSRAction3D}
\end{table*}

We utilize average pooling on predictions $\boldsymbol{Q}$ to obtain global semantics $\boldsymbol{q}_g\in\mathbb{R}^{c}$, and then duplicates it to $\boldsymbol{Q}_g\in\mathbb{R}^{M\times N^\prime \times c}$.
Moreover, we add 1D cosine positional encodings to $\boldsymbol{Q}_g$ to provide temporal clues, and the points in each frame share the same cosine encoding.
Finally, we put these updated embeddings into the decoder for spatiotemporal reconstruction.
We adopt FoldingNet \cite{yang2018foldingnet} as the decoder and exploit chamfer distance loss to optimize this self-supervised task as follows:
\begin{small}
\begin{equation}
    d(\boldsymbol{R},\boldsymbol{\hat P}_t^\prime)\!=\!\frac{1}{|\boldsymbol{R}|}\sum_{\hat x\in \boldsymbol{R}}\mathop{\min}_{p\in \boldsymbol{\hat P}_t^\prime}\|\hat x-p\|_2^2\!+\!\frac{1}{|\boldsymbol{\hat P}_t^\prime|}\sum_{p\in \boldsymbol{\hat P}_t^\prime}\mathop{\min}_{\hat x\in \boldsymbol{R}}\|p-\hat x\|_2^2,
\end{equation}
\end{small}

\noindent where $\boldsymbol{R}\in\mathbb{R}^{M\times N^\prime \times6}$ is reconstructed sequence.

Overall, the total loss with a regularized parameter $\lambda$ of our self-supervised framework consists of three parts:
\begin{equation}
    \mathcal{L}_{total} = \mathcal{L}_{l} + \mathcal{L}_{g} + \lambda d(\boldsymbol{R},\boldsymbol{\hat P}_t^\prime).
\end{equation}

\section{Experiments}
\subsection{Datasets}

We perform 3D action recognition on MSRAction3D and NTU-RGBD 60, and gesture recognition on NvGesture and SHREC'17 datasets.

\textbf{MSRAction3D} \cite{msr} includes 567 videos collected by Kinect, with a total of 23K frames and 20 categories performed by 10 subjects.
We use the same training and test splits as \cite{MeteorNet}.

\textbf{NTU-RGBD 60} \cite{ntu60} is collected by three cameras with different angles, containing 60 categories and 56880 videos performed by 40 subjects.
Cross-subject and cross-view evaluations are adopted.

\textbf{NvGesture} \cite{molchanov2016nvg} contains 1532 gesture videos focusing on touchless driver controlling, with a total of 25 classes.
We follow the previous work to split this dataset, where 1050 videos are used for training and 482 videos are for test \cite{min2020efficient}.

\textbf{SHREC'17} \cite{de2017shrec} collects 2800 videos performed by 28 subjects in two ways, \ie, using one finger or the whole hand.
These short videos generally contain dozens of frames and involve both coarse and fine gestures.
We adopt the same splits of training and test data as previous work \cite{min2020efficient}.

\subsection{Experimental Setting}

We perform pretraining on NTU-RGBD 60. 
Specifically, we consecutively sample 24 frames with stride 2, and each frame contains 1024 points.
This sequence is then divided into 6 segments.
Following \cite{pstnet}, the spatial search radius is 0.1, neighbors for the ball query is 9, and random scaling is adopted as data augmentation for our encoder.
Three transformer layers are utilized as our autoregressor.
We pretrain 200 epochs and set the batchsize to 88.
We use Adam optimizer and cosine annealing scheduler with the initial learning rate 0.0008.

Without special instructions, we adopt the pretrained point spatiotemporal encoder for downstream tasks, and add two linear layers with BN and ReLu for finetuning or one linear layer for linear evaluation.
We utilize SGD optimizer with momentum 0.9 and cosine scheduler with warmup 10 epochs.
The 16 batchsize corresponds to the 0.01 learning rate, and we follow the scale up rule.

\subsection{Comparison with State-of-the-art}

We perform extensive experiments and compare the performances with state-of-the-art supervised methods and other self-supervised methods.

\textbf{Transfer to MSRAction3D.} The finetune results are presented in Table \ref{MSRAction3D}, compared with skeleton-based, depth-based, and point-based methods.
We follow the settings of previous work \cite{MeteorNet} and test the performance under variable lengths.
Although the skeleton-based method Actionlet simultaneously feeds all frames, it is still exceeded by our CPR with 12 frames.
This shows that point cloud sequences contain richer spatiotemporal information.
Moreover, since 3D action recognition relies on temporal information, CPR obtains higher accuracy when inputting longer sequences.
Compared with other point-based supervised methods which train from scratch, CPR achieves improvements under diverse inputs, except that it is slightly lower than PST-Transformer on 24 frames.
This indicates that high-level semantics can be obtained by pretrained encoder and as a good initialization for transfer learning.
Compared with the self-supervised method that performs Recurrent Order Prediction task (ROP) \cite{minke}, CPR achieves higher accuracy.
ROP utilizes RGB values for pretraining and only focuses on clip-level temporal information, while our method utilizes local/global contrastive learning and reconstruction to explore richer spatiotemporal information just with points.

\begin{table}[ht]
    \centering
    \begin{tabular}{l|ccc}
    \toprule
    \textbf{Methods} &\textbf{Input} &\textbf{NvG} &\textbf{S17}\\
    \midrule
    Human\cite{molchanov2016nvg}                & RGB              & 88.4        & -   \\
    DG-STA~\cite{chenBMVC19dynamic}               & \small{skeleton}         & -           & 90.7 \\
    \midrule
    FlickerNet~\cite{flickernet}           & point            & 86.3        & -   \\
    PLSTM\small{-base}~\cite{min2020efficient}          & point            & 85.9        & 87.6 \\
    PLSTM\small{-early}~\cite{min2020efficient}         & point            & 87.9        & 93.5 \\
    PLSTM\small{-PSS}~\cite{min2020efficient}           & point            & 87.3        & 93.1 \\
    PLSTM\small{-middle}\cite{min2020efficient}        & point            & 86.9        & 94.7 \\
    PLSTM\small{-late}~\cite{min2020efficient}          & point            & 87.5        & 93.5 \\
    Kinet\cite{zhong2022nopain}                & point            & 89.1        & 95.2 \\
    PSTNet\cite{pstnet}                        & point            & 88.2        & 92.0 \\
    \midrule
    \textbf{CPR (Ours)}         & point            & 88.7        & 93.1 \\
    \bottomrule
    \end{tabular}
    \caption{Gesture recognition accuracy (\%) on NvGesture (NvG) and SHREC'17 (S17).}
    \label{NvGesture}
\end{table}

\textbf{Transfer to NvGesture and SHREC'17.} We perform self-supervised pretraining on human action datasets.
To further verify the generalization of our method, we also transfer the pretrained encoder to two gesture datasets.
During finetuning, the spatial search radius is 0.1, the neighbors of ball query is 9, and the frame number is 32.
CPR is compared with skeleton-based, RGB-based and other point-based supervised methods.
The results are presented in Table \ref{NvGesture}.
By finetuning with our self-supervised encoder, CPR facilitates the baseline PSTNet to produce comparable results.
This shows that the pretrained encoder has strong generalization ability, and exhibits powerful spatiotemporal modeling ability under diverse domains.

\textbf{Finetune on NTU-RGBD 60.} We keep the encoder and autoregressor for finetuning.
As shown in Table~\ref{NTU60}, we compare CPR with skeleton-based \cite{gcalstm,agclstm,asgcn,vafusion,2sagcn,dgnn}, depth-based \cite{mvdi}, and point-based supervised methods.
Particularly, Kinet builds a hierarchical motion branch, and 3DV explicitly encodes voxel-based motion cues.
Instead, without introducing hand-crafted motion features and complicated two-stream design, CPR achieves competitive accuracy under two evaluations.
This demonstrates the superiority of our self-supervised approach. 
By exploiting pretraining to fully mine motion information in the raw data, it can help the network acquire rich semantics without using additional dynamical model.
Our performance under cross-view evaluation is consistent with PSTNet++, and we will explore more advanced encoders in the future.

\begin{table}[ht]
    \centering
    \begin{tabular}{l|c|cccc}
    \toprule
    \multirow{2}{*}{\textbf{Methods}}    & \multirow{2}{*}{\textbf{Input} } & \multicolumn{2}{c}{\textbf{NTU-RGBD 60}}  \\
    & & Subject & View \\
    \midrule
    GCA-LSTM                    & skeleton          & 74.4 & 82.8  \\
    AGC-LSTM                    & skeleton          & 89.2 & 95.0  \\
    AS-GCN                      & skeleton          & 86.8 & 94.2  \\
    VA-fusion                   & skeleton          & 89.4 & 95.0  \\
    2s-AGCN                     & skeleton          & 88.5 & 95.1  \\
    DGNN                        & skeleton          & 89.9 & 96.1  \\
    \midrule
    MVDI                        & depth             & 84.6 & 87.3 \\
    \midrule
    PointNet++     & point             & 80.1 & 85.1  \\
    3DV (motion)                & voxel             & 84.5 & 95.4  \\
    3DV-PointNet++              & voxel + point     & 88.8 & 96.3  \\
    PSTNet                      & point             & 90.5 & 96.5  \\
    P4Transformer               & point             & 90.2 & 96.4  \\
    PSTNet++                    & point             & 91.4 & 96.7  \\
    PST-Transformer             & point             & 91.0 & 96.4 \\
    Kinet                       & point             & 92.3 & 96.4  \\
    \midrule
    \textbf{CPR (Ours)}         & point & 91.0  & 96.7  \\
    \bottomrule
    \end{tabular}
    \caption{Action recognition accuracy (\%) on the NTU-RGBD 60 dataset.}
    \label{NTU60}
\end{table}

\begin{table}[ht]
    \centering
    \begin{tabular}{l|cc|cc}
    \toprule
    \multirow{2}{*}{\textbf{SSL-Data}}  & \multicolumn{2}{c|}{\textbf{Linear}}  & \multicolumn{2}{c}{\textbf{Semi-supervision}}\\
          & \small{Data} & \small{Accuracy} &  \small{Data} & \small{Accuracy} \\
    \midrule
    MSR   & MSR  & 85.7    & 30\% MSR  & 86.4 \\
    NTU   & NTU  & 70.0    & 30\% NTU  & 83.3 \\
    \midrule
    NTU   & MSR  & 79.1    & 30\% MSR  & 87.2 \\
    \bottomrule
    \end{tabular}
    \caption{Recognition accuracy (\%) of linear evaluation and semi-supervised learning on MSRAction3D (MSR) and NTU-RGBD 60 (NTU). SSL-Data is self-supervised data.}
    \label{Semi-supervised}
\end{table}

\textbf{Linear and Semi-Supervised Learning.} To test whether the pretrained encoder has learned high-level semantics, we evaluate it on MSRAction3D and NTU-RGBD~60 under linear evaluation and limited training data.
Semi-supervised training data consists of randomly selected 30\% of the training samples from each class.
We conduct self-supervised pretraining on MSRAction3D and NTU-RGBD 60, respectively.
The results are shown in Table \ref{Semi-supervised}.
It is observed that the linear accuracy of \emph{MSR to MSR}, \emph{NTU to NTU}, and \emph{NTU to MSR} is 85.7\%, 70.0\%, and 79.1\%.
This shows that our self-supervised pretraining captures rich semantics beneficial for 3D action recognition.
After semi-supervision with 30\% data, all the performances get promoted compared with linear evaluations.
Since the 30\% MSRAction3D data is too small, the improvement of finetuning on MSR is limited.
In addition, the semi-supervised performance of \emph{NTU to MSR} is higher than that of \emph{MSR to MSR}.
This indicates that self-supervised pretraining with larger dataset can learn more general semantics while still needs limited data for transfer adaptation.

\subsection{Ablation Studies}

In this section, we show extensive ablation studies to explore that the length of whole input sequence utilized for our self-supervised framework, the design of point cloud sequences reconstruction, the effectiveness of hard negatives, and the influences of diverse self-supervised tasks.
These experiments are all pretrained on MSRAction3D and then finetuned on it.

\begin{table}[ht]
    \centering
    \begin{tabular}{c|ccccc}
    \toprule
    \diagbox{\textbf{SS-f}}{\textbf{Fin-f}} & 4 & 8 & 12 & 16 & 24 \\
    \midrule
    12   & 78.45 & 82.83 & 88.82 & 90.91 & 91.99  \\
    16   & 79.12 & 83.84 & 88.55 & 91.03 & 92.33  \\ 
    20   & 80.13 & 85.70 & 89.56 & 91.92 & 93.03  \\
    24   & 82.06 & 86.20 & 90.42 & 92.07 & \textbf{93.38}  \\
    \bottomrule
    \end{tabular}
    \caption{Finetune accuracy (\%) at different frames. Fin-f and SS-f are the frame number for finetune and self-supervised pretraining.}
     \label{length}
\end{table}

\begin{table}[ht]
    \centering
    \begin{tabular}{l|c|cc}
    \toprule
    \textbf{Reconstruction Target} & \textbf{Points of each frame} & \textbf{Acc.}\\
    \midrule
    One frame                      & 1024              &  91.32 \\
    \midrule
    One segment                    & 256               &  92.68  \\ 
    One segment + Color.     & 256               &  \textbf{93.38}  \\
    One segment + Color.     & 512               &  91.99 \\
    One segment + Color.     & 1024              &  92.36 \\
    \bottomrule
    \end{tabular}
    \caption{Finetune accuracy (\%) with different reconstruction targets. Color. means colorization.}
    \label{recontrust}
\end{table}

\begin{table}[ht]
    \centering
    \begin{tabular}{l|c}
    \toprule
    \textbf{Method}                            & \textbf{Acc.}\\
    \midrule
    Local contrastive prediction                    &  91.34  \\
    Global contrastive prediction                   &  90.89  \\
    Sequence reconstruction                         &  87.10   \\
    \midrule
    Local contrastive prediction with hard negatives   &  92.23 \\
    Local and global contrastive prediction         &  92.55 \\ 
    Contrastive prediction and reconstruction       &  \textbf{93.38} \\
    \bottomrule
    \end{tabular}
    \caption{Ablation studies on architecture design.}
    \label{task}
\end{table}

\textbf{How long of the point cloud sequence?} We perform self-supervised pretraining with variable input lengths and then finetune with diverse frames.
The results are shown in Table \ref{length}.
Notably, when sequence length of self-supervised pretraining is longer, it is more beneficial for finetuning under multiple frames. 
We finally choose 24 frames for self-supervised pretraining to cover as much dynamic information as possible.

\textbf{How to design reconstruction tasks?} For reconstructing branch, we try various experimental settings and the results are presented in Table \ref{recontrust}.
The accuracy of reconstructing one segment is higher than that of reconstructing one frame under the same number of points.
This may be due to the fact that spatiotemporal prediction is more challenging for self-supervised tasks. 
When applying point cloud sequence colorization, we achieve higher accuracy. 
Colorization is beneficial to distinguish diverse frames by assigning different timestamps for the target segment. 
We also try to reconstruct more points in each frame. 
However, this does not lead to improvements. 
It is possible that excessive raw points provide more low-level details and even noise, which does not help to promote generalization and discrimination.
Finally, we choose to reconstruct one colorized segment with 256 points in each frame.

\textbf{Why utilize hard negative samples?} The tokens of the former $S\!\!-\!\!2$ segments are temporally adjacent to the predictions and targets, and therefore they contain similar spatiotemporal semantics.
These tokens are mined as hard negatives to enhance local perception and discriminability.
The results in Table \ref{task} show that the accuracy of local contrastive prediction without hard negatives is 91.34\%, and the accuracy increases to 92.23\% after adding them. 
This indicates that mining hard negative samples is crucial for the design of self-supervised tasks.

\textbf{How the effects of each task?} We respectively evaluate the effectiveness of local contrastive prediction, global contrastive prediction, and point cloud sequence reconstruction tasks. 
The results are shown in Table \ref{task}.
The hard negatives can increase the performance of local contrastive prediction by about 1\%.
Clearly, by introducing the reconstruction task, the finetune performance has increased by 0.83\%, which indicates reconstructing raw inputs helps to learn semantics information suitable for 3D action recognition.
Compared with only utilizing local contrastive prediction, the introduction of global contrastive prediction increases the accuracy by 0.32\%.
This shows that global discrimination and local perception are all essential for capturing multi-granularity representations.
Overall, the most beneficial paradigm is to combine contrastive prediction and point cloud sequence reconstruction.

\begin{figure}[t]
\centering
\setlength{\abovecaptionskip}{0.4cm}
\includegraphics[width=0.9\linewidth]{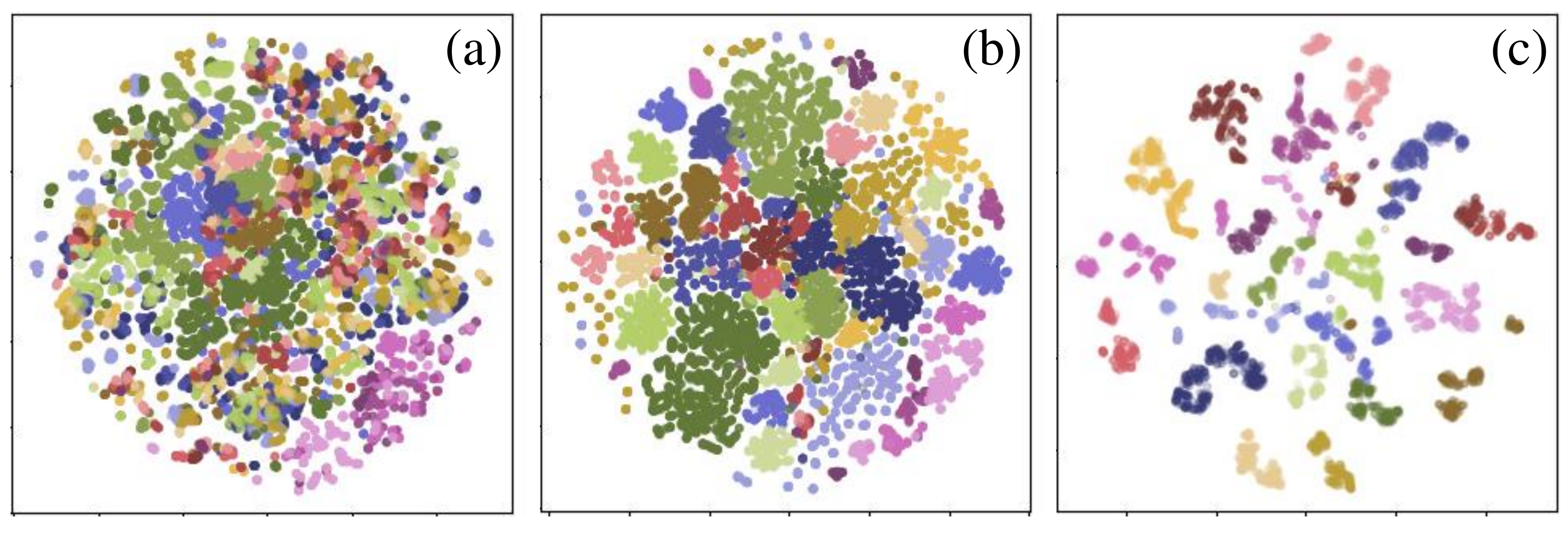}
\caption{Visualization of the t-SNE features. We visualize the feature distributions of our self-supervised encoder (a) after pretraining, (b) after finetuning on NTU-RGBD 60, and (c) after finetuning on MSRAction3D.}
\label{vis}
\end{figure}

\subsection{Visualization}

In Figure \ref{vis}, we visualize the feature distributions of t-SNE after pretraining (a) and after finetuning (b)(c).
It can be seen that after pretraining, there are approximate boundaries between 60 categories.
This illustrates that self-supervised pretraining can learn certain high-level semantics.
From Figure \ref{vis}(c), it is observed that each cluster has clear outlines, indicating that the representations learned by our self-supervised framework have strong generalization and transferability.
Moreover, the proposed framework can guide the encoder to obtain domain-agnostic and general knowledge.

\section{Conclusion}

In this work, we propose a unified contrastive prediction and reconstruction self-supervised framework for dynamic point cloud understanding. 
By integrating discriminative and generative self-supervised tasks, it makes the learned representations with both powerful instance discrimination and local perception.
Extensive experiments under linear evaluation, semi-supervised learning, and transfer learning are performed on multiple benchmarks to demonstrate the superiority of our self-supervised framework. 
In the future, we will explore more dynamic point cloud related downstream tasks.

\section{Acknowledgments}
This paper is sponsored by National Natural Science Foundation of China (No.61673270; No.61973212).

\bibliography{aaai23}

\end{document}